\documentclass[runningheads]{llncs}

\usepackage{eccv}

\usepackage{eccvabbrv}

\usepackage{graphicx}
\usepackage{booktabs}

\usepackage[accsupp]{axessibility}  

\usepackage[pagebackref,breaklinks,colorlinks,citecolor=eccvblue]{hyperref}

\usepackage{orcidlink}

\begin{document}

\title{DDRN:a Data Distribution Reconstruction Network for Occluded Person Re-Identification} 

\titlerunning{Data Distribution Reconstruction Network}

\author{Zhaoyong Wang\inst{1} \and
Yujie Liu\inst{1,2}\and
Mingyue Li\inst{1} \and   
Wenxin Zhang\inst{1} \and
Zongmin Li\inst{1}}



\institute{China University of Petroleum Huadong - Qingdao Campus, Qingdao, 266580,
China.  \and
Corresponding author: 
\email{liuyujie@upc.edu.cn;
}}

\maketitle

\begin{abstract}
In occluded person re-identification(ReID), severe occlusions lead to a significant amount of irrelevant information that hinders the accurate identification of individuals. These irrelevant cues primarily stem from background interference and occluding interference, adversely affecting the final retrieval results. Traditional discriminative models, which rely on the specific content and positions of the images, often misclassify in cases of occlusion. To address these limitations, we propose the Data Distribution Reconstruction Network (DDRN), a generative model that leverages data distribution to filter out irrelevant details, enhancing overall feature perception ability and reducing irrelevant feature interference. Additionally, severe occlusions lead to the complexity of the feature space. To effectively handle this, we design a multi-center approach through the proposed Hierarchical SubcenterArcface (HS-Arcface) loss function, which can better approximate complex feature spaces. On the Occluded-Duke dataset, we achieved a mAP of 62.4\% (+1.1\%) and a rank-1 accuracy of 71.3\% (+0.6\%), surpassing the latest state-of-the-art methods(FRT) significantly.
\end{abstract}

\section{Introduction}
\label{sec:intro}

Person ReID is a task that aims to identify individuals across different cameras, considering various factors such as viewpoints, lighting conditions, and locations. It finds applications in various fields, especially in intelligent security\cite{eom2019learning,ye2021channel,zhai2020ad,zheng2019pyramidal}, which has led to its widespread attention. With the advent of deep learning, ReID methods based on deep learning have become the mainstream\cite{zheng2015scalable,zheng2017unlabeled,li2014deepreid,wei2018person}.

However, the occlusion problem continues to pose a significant challenge in Person ReID, as images are frequently captured in complex environments or crowded scenarios. Severe occlusions lead to a significant amount of irrelevant information that hinders the accurate identification of individuals. Despite the attention the task has received, fully addressing the occlusion problem is still an ongoing research area, and many works continue to explore methods to overcome occluded ReID challenges\cite{zhang2020semantic,huang2020human,gao2020pose,wang2020high}.

\begin{figure}[!t]
  \centering
  \includegraphics[width=1.0\linewidth]{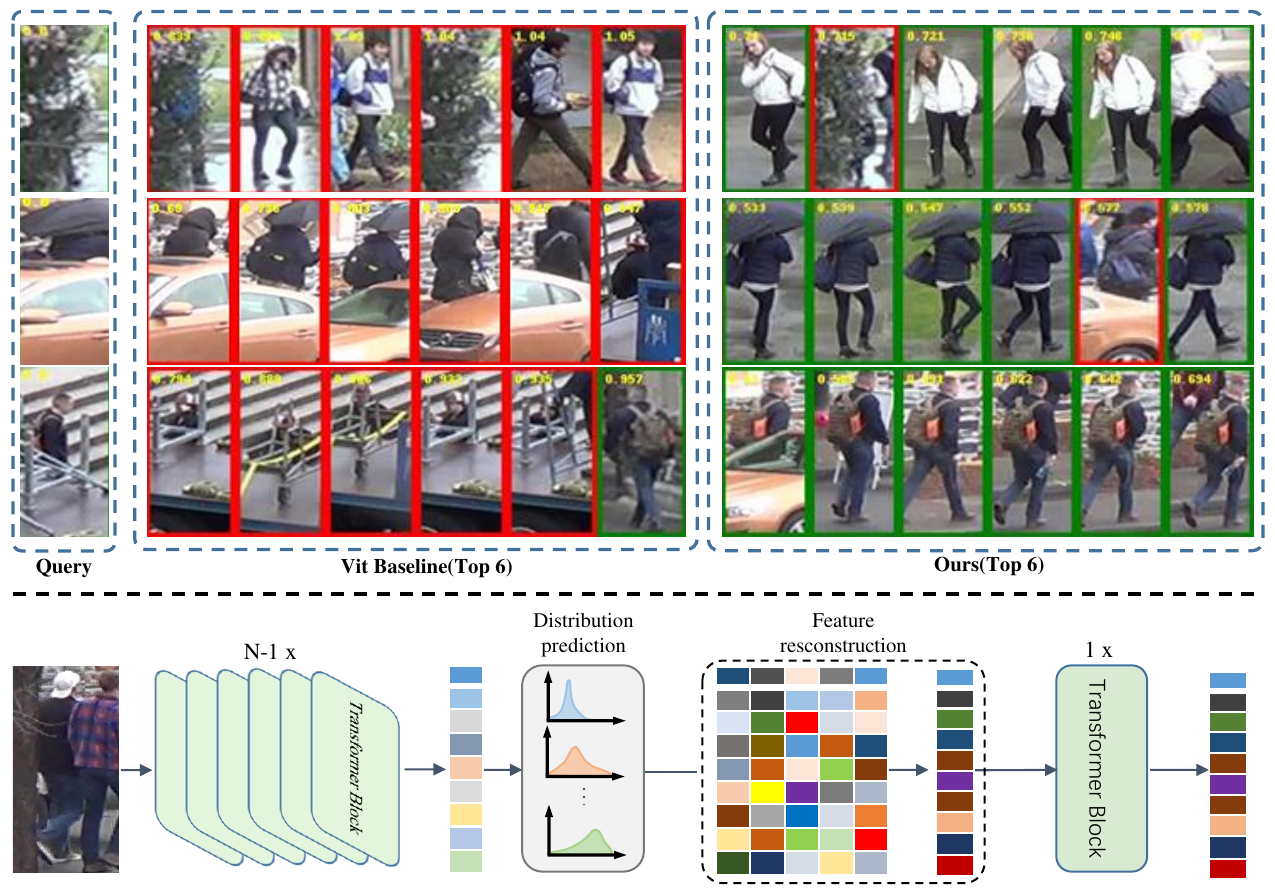}
  \caption{Retrieval results (top) for both the Vit Baseline and our approach, along with the overall structure (bottom) of our proposed method. The upper part of the figure illustrates the impact of occlusion and background interference on retrieval results. The green boxes represent correct retrieval results, while the red boxes represent incorrect retrievals. The leftmost image represents the query image, the middle column shows the retrieval results of the Vit Baseline, and the rightmost column displays the retrieval results of our proposed method. The lower part presents an overview of our proposed solution, which achieves feature reconstruction by predicting the distribution of the intermediate process.}
  \label{fig:1}
\end{figure}

Based on our analysis and investigation, we have identified two main problems in occluded person ReID:

\begin{enumerate}
\item Occlusion interference: As shown in the upper part of Figure \ref{fig:1}, different occlusions may affect the final features during network inference, leading to interference and inaccurate retrieval results.
\item Background interference: As shown in the last two rows of the upper part of Figure \ref{fig:1}, different images may have the same background, resulting in images with different identities being erroneously retrieved as the same identity.
\end{enumerate}

The two challenges mentioned above are clearly evident in the retrieval results, highlighting the necessity of finding more effective solutions. Previous research has made specific progress in this field, which can be broadly classified into two categories. In the context of addressing these challenges, some approaches adopt external network models to guide the network's focus towards the most discriminative parts\cite{miao2019pose, wang2020high, tan2022mhsa, huang2020human}. Meanwhile, another set of strategies employs attention mechanisms to selectively direct the network's attention to the most discriminative parts\cite{guo2022attention, wang2022feature, tay2019aanet, zhang2020relation, li2018harmonious}. However, both approaches fail to fully address the challenges in scenarios where occlusions are severe, as depicted in the first row of the upper part of Figure \ref{fig:1}. Specifically, external network fail to perform effectively, and attention mechanisms may still introduce some background information into the final features, leading to interference with the retrieval results.

The mentioned approaches belong to discriminative models and rely on the specific content and positions of the images, making it challenging to differentiate them in the feature space. To better eliminate the occlusion and background interference issues mentioned earlier, we propose our generative model, Data Distribution Reconstruction Network(DDRN). Unlike the two types of networks mentioned earlier, we employ a feature space reconstruction approach by predicting the mean and variance of the intermediate process in the network, instead of relying on the positions of different parts in the images. The network architecture of DDRN is illustrated in the lower part of Figure \ref{fig:1}. The retrieval results of DDRN are shown in the rightmost part of Figure \ref{fig:1}, demonstrating its effectiveness in tackling the two types of problems.

Predicting the mean and variance of features directly entails predicting a continuous distribution, which is then converted into discrete values through sampling. However, the input data itself is discrete. Therefore, the features undergo a transformation from discrete to continuous and back to discrete values, which undoubtedly increases the difficulty of network training. In light of this, DDRN aims to directly predict discrete distributions, reducing the complexity of network training and increasing its generality. So DDRN involves discrete values instead of continuous ones, we introduce an Embedding Space\cite{van2017neural} comprising a series of discrete vectors. We apply an orthogonal loss to ensure that these discrete vectors span the entire feature space, enforcing pairwise standard orthogonality in the Embedding Space.

In occluded ReID scenarios, some images suffer from severe occlusions, making it challenging to distinguish them correctly. SubcenterArcFace loss\cite{deng2020sub} can effectively prevent these images from interfering with the network training process and enhance the network's generalization ability. However, it may result in reduced compactness within each class, leading to an increased feature space occupied by each category during training and larger average distances between training samples within the same class. To address this issue, we propose the Hierarchical SubcenterArcface(HS-Arcface) loss, which can uniformly spread multiple feature distributions. 

Our main contributions to this paper are summarized as follows:

\begin{enumerate}
\item Pioneering generative networks for occluded person ReID, DDRN introduces the Embedding Space for feature reconstruction.\ref{sec:embedding space}

\item Proposing an orthogonal loss promotes uniform feature distribution in the Embedding Space.\ref{sec:orthogonal loss}

\item Introducing the HS-Arcface loss overcomes challenges in occluded person ReID, improving feature discriminative power.\ref{sec:hs-arcface loss}

\item Experiments on common datasets consistently achieve state-of-the-art results, showcasing superior accuracy and robustness in handling occlusion challenges.\ref{sec:experiments}
\end{enumerate}

\section{Related Works}
This section provides a brief overview of existing methods for holistic person ReID and occluded person ReID.

\subsection{Holistic Person ReID}
Person ReID has made remarkable progress in recent years, mainly driven by the availability of large datasets and advancements in powerful GPUs. Person ReID are categorized into three main types: global feature representation learning\cite{wu2016personnet,li2018harmonious,yang2019towards,he2021transreid}, local feature representation learning\cite{li2017learning,su2017pose,sun2018beyond}, and auxiliary feature representation learning\cite{wang2018transferable,chang2018multi,sarfraz2018pose,shen2018person}.

Several notable works have contributed to the development of person ReID. For instance, Luo \textit{et al}.\cite{luo2019bag} introduced the BN-Neck structure in the CNN-based ReID framework, providing a strong baseline for holistic person ReID. Zhang \textit{et al}.\cite{zhang2017alignedreid} proposed an automatic part feature alignment method using shortest path loss without requiring additional supervision or explicit posture information. Sun \textit{et al}.\cite{sun2018beyond} developed a general parts-level feature learning method adaptable to various parts-partitioning strategies. Additionally, attentional mechanisms have been incorporated to ensure that the model focuses on human regions and extracts more effective features\cite{tay2019aanet,li2018harmonious}. Jia \textit{et al}. \cite{jia2022learning} presented DRL-Net (Disentangled Representation Learning Network), a model that doesn't necessitate strict alignment and extra supervision. For local feature representation or aligning partial features within the transformer, Zhu \textit{et al}. \cite{zhu2023aaformer} introduced the first alignment-based Transformer ReID framework, known as AAformer. 

Despite the advancements, the methods above have limitations in accurately retrieving people under occlusion, which restricts their application, particularly in occluded scenarios.

\subsection{Occluded Person ReID}
The occluded person ReID was initially proposed by Zhuo \textit{et al.} \cite{zhuo2018occluded}, and subsequent research has aimed to tackle this problem. Miao \textit{et al.} \cite{miao2019pose} integrated pose estimation into their PGFA framework to detect non-occluded areas, while Ren \textit{et al.} \cite{ren2020semantic} developed the SGSFA method, leveraging attention to enhance local feature representation.

Other approaches introduced auxiliary information to better distinguish between pedestrian and non-pedestrian regions. \textit{Wang et al.} \cite{wang2020high} proposed the HOReID framework, which employs a graph neural network to model topology information for improved differentiation between pedestrian and non-pedestrian regions. Gao \textit{et al.} \cite{gao2020pose} introduced visibility prediction through self-supervision in their PVPM framework. Huang \textit{et al.} \cite{huang2020human} implemented visible area detection in conjunction with person parsing in their HPNet, and Zhang \textit{et al.} \cite{zhang2020semantic} designed the SOAR method to extract features from local, global, and semantic perspectives. BPBreID\cite{somers2023body} designed two modules for predicting body part attention maps and producing body part-based features of the ReID target using identity and coarse human parsing labels.

More recently, Tan \textit{et al.} \cite{tan2022mhsa} proposed MHSA-Net, a framework that combines multiple attention mechanisms, while Jia \textit{et al.} \cite{jia2021matching} introduced the MoS framework, which utilizes set matching to optimize the feature measure. Finally, He \textit{et al.} \cite{he2021transreid} presented TransReID, the first pure transform-based architecture for person ReID. Additionally, LDS\cite{zang2021learning} employed several self-supervised operations to simulate various challenging problems and employs distinct networks to address each challenging problem.Tan \textit{et al.}\cite{tan2022dynamic} proposed a novel Dynamic Prototype Mask(DPM) based on two self-evident prior knowledge, which masks the regions with occlusions, thereby aligning occluded pedestrians with complete pedestrians.

In contrast to the methods mentioned above, DDRN is a generative model. It predicts the data distribution in the feature space and reconstructs the features to eliminate interference caused by occlusions and background. By leveraging the learned distribution information, DDRN effectively addresses the challenges of occluded person ReID, resulting in improved accuracy and robustness in occlusions and background interference.

\begin{figure*}
  \centering
  \includegraphics[width=1.0\linewidth]{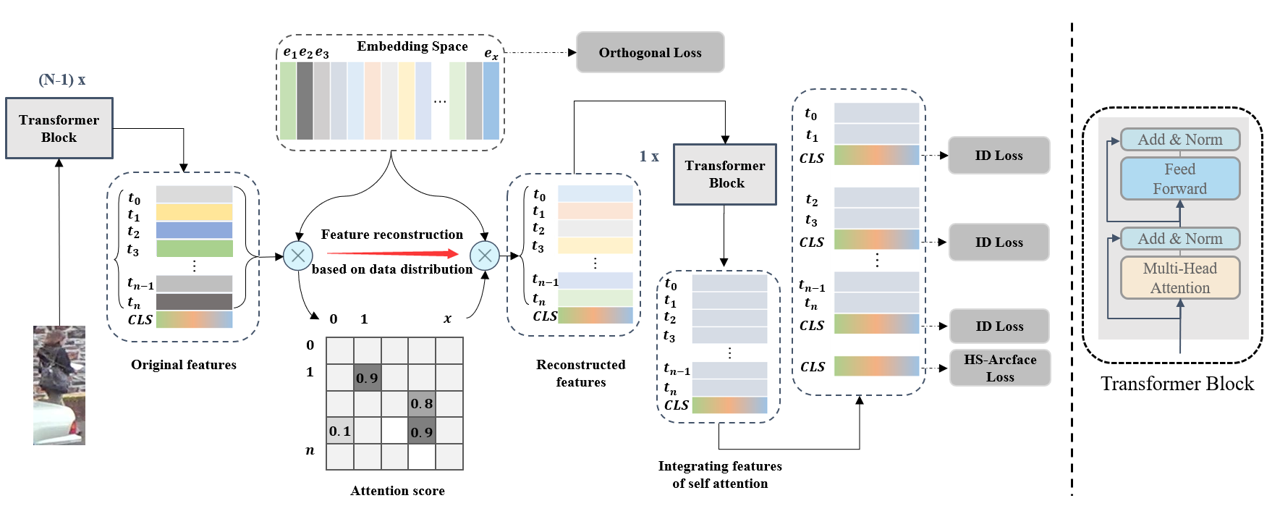}
  \caption{The network architecture of DDRN. It using the standard transformer block as feature extractor. The features between the $N-1_{th}$ and $N_{th}$ layers are replaced by the most similar vectors in the Embedding Space. Orthogonal Loss is employed to ensure that the vectors in the Embedding Space can represent different types of features separately. The final global CLS token utilizes our proposed HS-Arcface loss. the CLS token, which fuses local features, adapts ID loss.}
  \label{fig:2}
\end{figure*}

\section{Data Distribution Reconstruction Network}

\subsection{Overall Framework}

 Unlike traditional discriminative networks, which rely on specific image positions, DDRN adopts a generative model, predicting image distribution and reconstructing features. This mitigates occlusion and background interference, improving identification accuracy.

DDRN's architecture (Figure \ref{fig:2}) follows an encoder-decoder structure with N Transformer Blocks\cite{dosovitskiy2020image} and an Embedding Space\cite{van2017neural}. After serialization and Position Embeddings\cite{dosovitskiy2020image}, features undergo cross-attention computations and are reconstructed using the $N_{th}$ Transformer Block. 

To further enhance the network's performance, we apply orthogonalization to the vectors in the Embedding Space. Orthogonalization of Embedding Space vectors enhances discriminative spatial projection, reducing redundancy for improved generalization. The proposed HS-Arcface constrains global CLS token\cite{dosovitskiy2020image}, while standard ID Loss constrains CLS token with local information. Triplet losses, computed with corresponding Bottle Neck layers, contribute to DDRN's effectiveness in handling occlusion and background interference in occluded person ReID.

\subsection{Transformer Architecture}

In our DDRN network (Figure \ref{fig:2}), we adapt the standard Vision Transformer architecture by incorporating feature reconstruction between the $N-1_{th}$ and $N_{th}$ layers. This unique modification allows us to preserve the global attention mechanism of the Transformer while reconstructing image features. The decoder corresponds to the $N_{th}$ Transformer Block, and the encoder spans the $1_{st}$ to $N-1_{th}$ Transformer Blocks. This arrangement effectively addresses occlusion and background interference challenges in retrieval results.

Additionally, our approach ensures that the CLS token shares weights throughout the Transformer, maintaining contextual and global attention. Other tokens are replaced with vectors from the Embedding Space. After the final Transformer Block, global attention is applied to integrate the reconstructed feature representations into the overall contextual information. This design enhances DDRN's ability to handle occlusion and background interference in occluded person ReID.

\subsection{Embedding Space}
\label{sec:embedding space}
We adopt a feature-level data reconstruction approach inspired by VAE\cite{kingma2013auto} and GAN\cite{goodfellow2020generative} to handle occlusion and background interference in person ReID. Our objective is to predict the mean and variance of input features, eliminating irrelevant information for accurate identification and enhancing network robustness. To address slow convergence in direct predictions using VAE or GAN models, we propose training a discrete Embedding Space.

The Embedding Space, comprising a set of embeddings, replaces specific features with the nearest embedding during acquisition, simulating mean and variance predictions. To overcome gradient truncation issues with Argmax, we adopt the default method\cite{jang2016categorical} of copying decoder gradients to the encoder.

Despite this, gradient inconsistencies persist, prompting the use of Gumbel Softmax\cite{jang2016categorical} for resampling. During the forward pass, Argmax is employed, while Gumbel Softmax is used for backpropagation in the backward pass. The resampling process can be described as follows:
\begin{equation}
    t'_i = \sum_{j=1}^{x} (\mathcal{G}{(\mathcal{S}(t_i \times E^T))})_j \times E_j 
\end{equation}

Here, $\mathcal{S}$ denotes Softmax, $\mathcal{G}$ represents Gumbel Softmax, and $E$ represents the Embedding Space. $x$ represents the number of vectors in the Embedding Space, and $t_i$ denotes the $i_{th}$ vector in the Embedding Space. $\mathcal{G}$ computes the Gumbel Softmax as follows:
\begin{equation}
\mathcal{G}{(v_i)} = \frac{e^{{(v_i+G_i)}/\tau}}{\sum_{j=1}^{x}e^{{(v_j+G_j)}/\tau}} 
\end{equation}
where $G_i$ follows the Gumbel distribution. The temperature coefficient $\tau$ determines the degree of approximation to the one-hot distribution as it decreases.

However, the Gumbel-Softmax method only makes the differentiable part of the output approach a one-hot vector as the temperature coefficient changes, but it cannot generate an actual one-hot vector on its own. Thus, an argmax function is still required to convert it into a one-hot representation. To achieve this, we employ the Gumbel-Softmax in conjunction with the Straight-Through Estimator\cite{bengio2013estimating}, allowing us to effectively address the challenge of gradient truncation during training and improve the performance of our DDRN model in handling occluded person ReID tasks.

\begin{figure*}[!ht]
  \centering
  \includegraphics[width=1.0\linewidth]{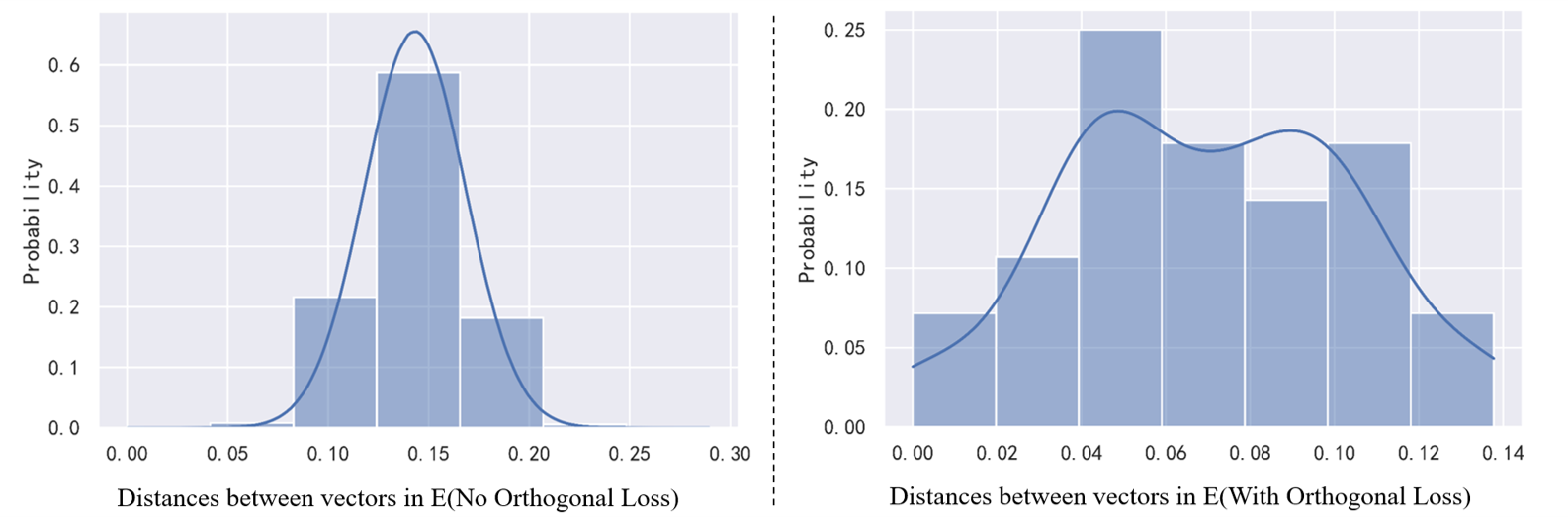}
  \caption{The vectors in the Embedding Space have cosine distances between each other, except for themselves. To demonstrate the proposed Orthogonal Loss, we statistically analyze their cosine distance distribution. The x-axis represents the cosine distance, and the y-axis represents the corresponding probabilities. The curve in the plot is obtained using kernel density estimation with the Gaussian kernel function.}
  \label{fig:3}
\end{figure*}

\subsection{Orthogonal Loss}

\label{sec:orthogonal loss}
We introduce the orthogonal loss to enhance diversity and eliminate redundancy in the Embedding Space. This loss aims to create a feature representation with broad and distinct characteristics, preventing redundancy. By encouraging orthogonality among embedding vectors, the loss promotes a diverse and non-collinear feature representation. The orthogonal loss is calculated using the following formula:
\begin{equation}
    \mathcal{L}_{Orthogonal} = \lVert \ddot{E}\ddot{E}^T - \mathcal{I}_N \rVert_F
\end{equation}
where $\lVert \cdot \rVert_F$ denotes the Frobenius norm, and $\mathcal{I}_N \in N \times N$ represents the identity matrix of size $N$. The matrix $\ddot{E}$ is obtained by normalizing the $E$ matrix, ensuring that each row is L2 normalized.

Incorporating the orthogonal loss during training encourages embedding vectors to be independent and non-redundant, creating a more expressive Embedding Space. This helps the model capture diverse features and reduces overfitting. Figure \ref{fig:3} demonstrates that employing the orthogonal loss increases the cosine distance between embedding vectors, manifesting in cosine values approaching 0 between vectors. Overall, the orthogonal loss enhances DDRN's robustness and generalization, enabling superior performance in occluded person ReID by effectively handling challenging conditions.

\subsection{HS-Arcface Loss}
\label{sec:hs-arcface loss}
In occluded person ReID, conventional cross-entropy loss lacks constraints on intra-class distances, which cannot extract robust pedestrian representations, affecting performance in occluded scenarios. 

\begin{figure}
  \centering
  \includegraphics[width=1.0\linewidth]{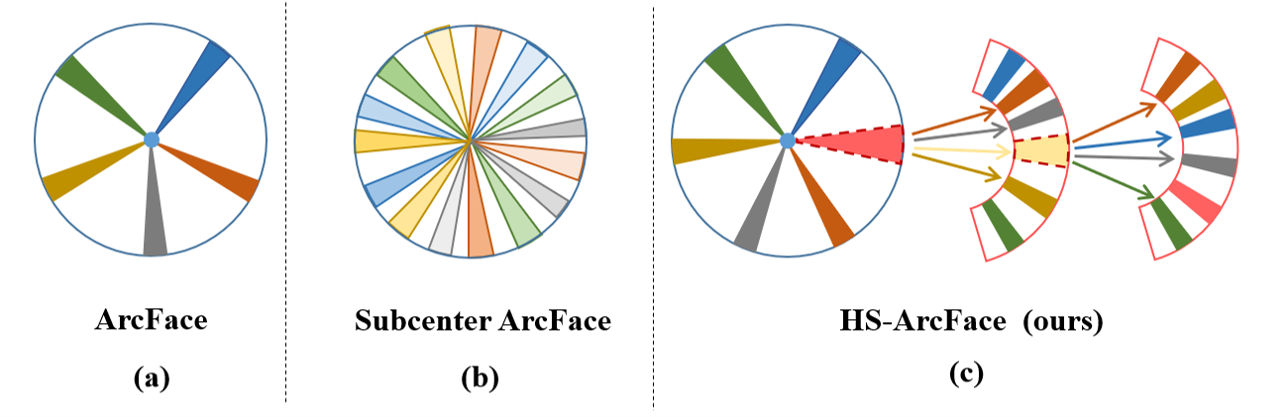}
  \caption{Comparison among ArcFace, Subcenter ArcFace, and HS-ArcFace. We use the disk to represent the entire feature space, where ArcFace(a) performs classification in the feature space. SubCenterArcFace(b) introduces multiple sub-center representations on top of ArcFace. HS-Arcface(c) addresses the reduced inter-class distances caused by introducing k sub-center representations in SubCenterArcFace. }
  \label{fig:5}
\end{figure}

ArcFace, as shown in Figure \ref{fig:5} (a), introducing a margin, improves intra-class and inter-class distances but faces challenges in severe occlusions. SubCenterArcface enhances generalization by introducing k class centers, reducing distances between different class centers, as shown in Figure \ref{fig:5} (b). On this basis, we propose Hierarchical SubcenterArcface (HS-Arcface) introducing an extra clustering center for appropriate inter-class distances. As depicted in Figure \ref{fig:5} (c), HS-Arcface introduces an additional clustering center as a marker for the following layer classification, ensuring appropriate inter-class distances. When k=1, HS-Arcface degenerates into ArcFace. 

 \begin{figure*}
  \centering
  \includegraphics[width=1.0\linewidth]{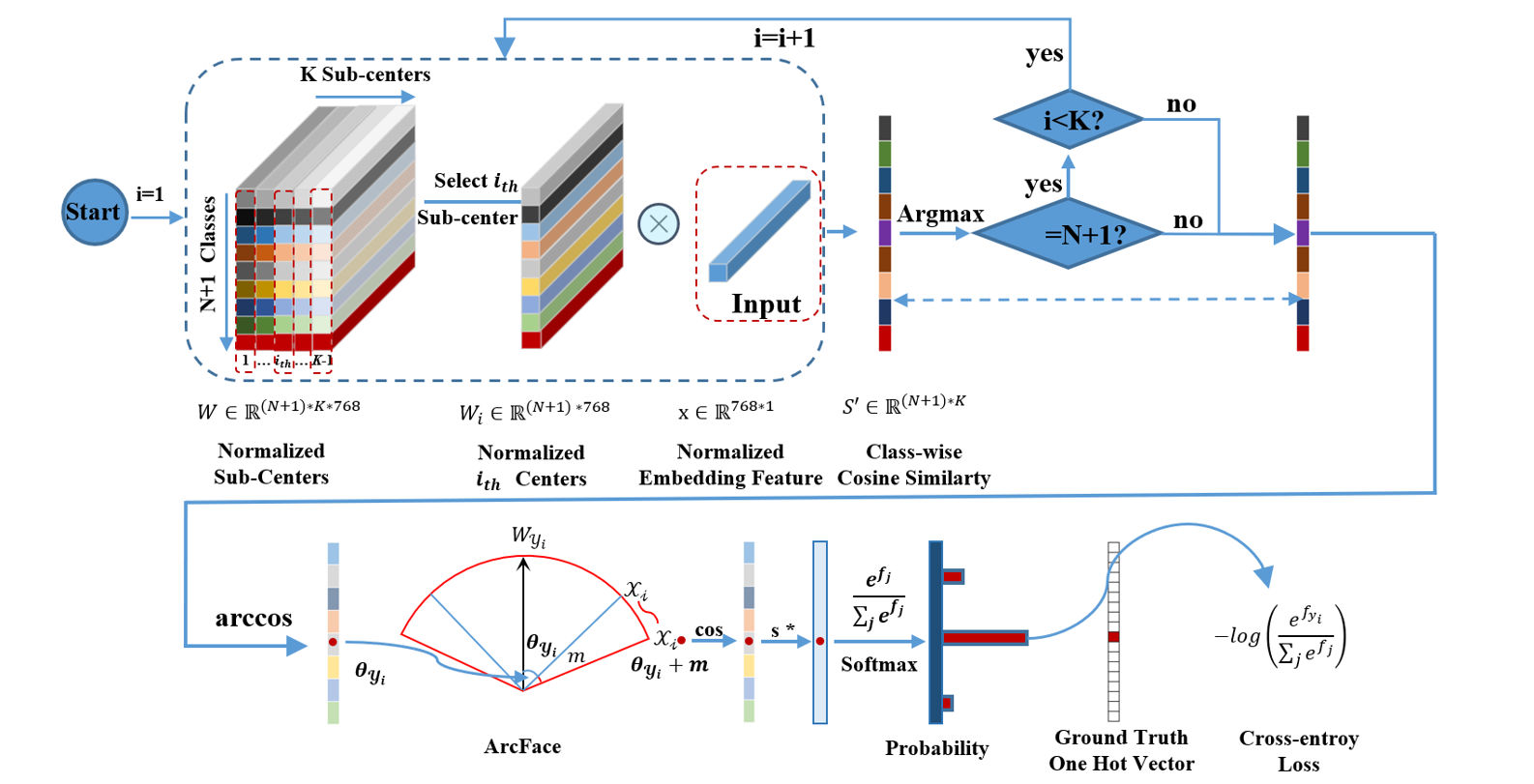}
  \caption{The network architecture of HS-Arcface. In the initial stage of HS-Arcface, we set $i$ to 1. We select the $i_{th}$ sub-center among the K sub-center representations and compute its similarity with the input feature. Then, we determine its corresponding class label. If the class label equals N+1, we consider the need for further classification at the next level. So HS-Arcface introduces an additional clustering
center(N+1) as a marker for the following layer classification. If $i$ is less than K, it means we have not reached the final layer, so we increment $i$ by one and continue the loop. Finally, we apply the ArcFace operation to the output.}
  \label{fig:4}
\end{figure*} 

 In comparison to Subcenter Arcface, HS-Arcface introduces an additional clustering center along with its subcenters. Figure \ref{fig:4} illustrates the implementation of HS-Arcface. Samples falling into this extra clustering center proceed to the next layer for classification. After determining $\theta$, the computation of HS-Arcface is formulated as Equation \ref{eq:4}. 
\newcounter{TempEqCnt} 
\setcounter{TempEqCnt}{\value{equation}} 
\setcounter{equation}{3} 

\begin{equation}
\mathcal{L}_{\text{HS-Arcface}} = -\frac{1}{N}\sum_{i=1}^{N}\log\left(\frac{e^{s(\cos(\theta_{y_i})+ m)}}{e^{s(\cos(\theta_{y_i})+ m)} + \sum_{j\neq y_i}e^{s\cos(\theta_j)}}\right)
\label{eq:4}
\end{equation}

Here, $N$ represents the batch size, $y_i$ denotes the accurate label of the $i$-th sample, $\theta_{y_i}$ represents the actual angle of the $i$-th sample, $\theta_j$ denotes the angle of the $j$-th sample, $m$ is the hyperparameter margin, and $s$ is the scaling parameter. The HS-Arcface loss function effectively captures the essential features and promotes better discrimination between different classes, ultimately leading to improved occluded person ReID performance.

\section{Experiments}
\label{sec:experiments}
\subsection{Datasets and Evaluation Setting}

\textbf{Occluded-DukeMTMC}\cite{miao2019pose} is the largest and most challenging dataset for occluded person ReID, containing 15,618 training images, 17,661 gallery images, and 2,210 blocked query images.

\textbf{Occluded REID}\cite{zhuo2018occluded} includes 2,000 images of 200 occluded persons, with each identity having 5 full-body person images and 5 occluded person images with different types of occlusion.

\textbf{Partial REID}\cite{zheng2015partial} is a dataset that includes 600 images of 60 people, with 5 full-body images and 5 occluded images per person. These images are collected on campus by 6 cameras from different viewpoints, backgrounds, and different types of occlusion.

\textbf{Market-1501}\cite{zheng2015scalable} consists of 32,668 images of 1,501 persons. 750 identities are utilized for training, and the remaining 751 identities are used for testing. The dataset includes 3,368 query images as a probe set and 19,732 images as a gallery set.

\textbf{DukeMTMC-reID}\cite{zheng2017unlabeled} contains 16,522 training images of 702 identities, 2,228 query images of the other 702 identities, and 17,661 gallery images.

\textbf{Evaluation Protocol:} To ensure a fair comparison with existing methods, all experiments follow the standard evaluation settings in person ReID research. The Cumulative Matching Characteristic (CMC) and mean Average Precision (mAP) are adopted as performance evaluation metrics. All experiments are conducted in the single query setting.

\subsection{Implementation Details}
We use standard Transformer size (N=12) and apply image preprocessing with strategies like Padding, Crop, Resize, and Random Erasing\cite{zhong2020random}. Training involves a Transformer stride size of 11, Batch Size of 64, and 120 epochs.

\begin{table}[!ht]
\centering  
\caption{Performance comparison with state-of-the-art methods on Occlude-DukeMTMC(O-Duke), Occluded-ReID(O-REID), Partial-REID(P-REID), Market-1501 and DukeMTMC datasets.}  
\setlength{\tabcolsep}{1.0mm}{
    \begin{tabular}{lcccccccccc}
    \toprule[1pt] 
           & \multicolumn{2}{c}{O-Duke} & \multicolumn{2}{c}{O-REID} & \multicolumn{2}{c}{P-REID} & \multicolumn{2}{c}{Market-1501}  & \multicolumn{2}{c}{DukeMTMC}\\
    Method & R@1 & mAP & R@1 & mAP & R@1 & mAP & R@1 & mAP & R@1 & mAP\\ 
    \midrule[0.5pt]  
    \emph{i) CNN} & & & & & & \\
    PCB\cite{sun2018beyond} & 42.6 & 33.7 & 41.3 & 38.9 & 66.3 & 63.8 & 92.3 & 77.4 & 81.8 & 66.1\\ 
    RE\cite{zhong2020random} & 40.5 & 30.0 & - & - & 54.3 & 54.4 & - & - & - & -\\
    DSR\cite{he2018deep} & 40.8 & 30.4 & 72.8 & 62.8 & 73.7 & 68.07 & - & - & - & - \\
    FD-GAN\cite{ge2018fd} & 40.8 & - & - & - & - & - & - & - & - & -\\
    SFR\cite{he2018recognizing} & 42.3 & 32 & - & - & 56.9 & - & - & - & - & -\\
    FRR\cite{he2019foreground} & - & - & 78.3 & 68.0 & 81.0 & 76.6 & - & - & - & -\\
    PVPM\cite{gao2020pose} & 47 & 37.7 & 70.4 & 61.2 & - & - & - & - & - & - \\
    SSGR\cite{yan2021occluded} & 65.8 & 57.2 & 78.5 & 72.9 & - & - & - & - & - & -\\
    LDS\cite{zang2021learning} & 64.3 & 55.7 & - & - & - & -& - & - & - & -\\
    BPBreID\cite{somers2023body} & \textbf{71.3} & 57.5 & 77.0 & 70.9 & - & - & - & - & - & -\\
    SPReID\cite{kalayeh2018human} & - & - & - & - & - & - & 93.7 & 83.4 & 86.0 & 73.3 \\
    VPM\cite{sun2019perceive} & - & - & - & - & - & - & 93.0 & 80.8 & 83.6 & 72.6 \\
    MGN\cite{wang2018learning}  & - & - & - & - & - & - & 95.7 & 86.9 & 88.7 & 78.4 \\
    STNReID\cite{luo2020stnreid} & - & - & - & - & - & - & 93.8 & 84.9 & - & - \\
    FIDI\cite{yan2021beyond} & - & - & - & - & - & - & 94.5 & 86.8 & 88.1 & 77.5 \\
    CAL\cite{rao2021counterfactual} & - & - & - & - & - & - & 94.5 & 87.0 & 87.2 & 76.4 \\
    CDNet\cite{li2021combined} & - & - & - & - & - & -& 95.1 & 86.0 & 88.6 & 76.8\\
    AAformer\cite{zhu2023aaformer} & - & - & - & - & - & -& 95.4 & 87.7 & 90.1 & 80.0\\ 
    PGFA\cite{miao2019pose} & 51.4 & 37.3 & - & - & 69.0 & 61.5 & 91.2 & 76.8 & 82.6 & 65.5\\
    HOReID\cite{wang2020high} & 55.1 & 43.8 & 80.3 & 70.2 & 85.3 & - & 94.2 & 84.9 & 86.9 & 75.6 \\
    OAMN\cite{chen2021occlude} & 62.6 & 46.1 & - & - & 86.0 & - & 93.2 & 79.8 & 86.3 & 72.6\\
    ISP\cite{zhu2020identity} & 62.8 & 52.3 & - & - & - & - & 95.3 & \textbf{88.6} & 89.6 & 80.0\\
    Pirt\cite{2021MaPirt} & 60.0 & 50.9 & - & - & - & - & 94.1 & 86.3 & 88.9 & 77.6\\
    \midrule[0.5pt] 
    \emph{ii) transformer} & & & & & & \\
    Vit Baseline & 61.1 & 53.1 & 81.2 & 76.7 & 73.3 & 74.0  & 94.7 & 86.8 & 88.8 & 79.3\\
    PAT\cite{li2021diverse} & 64.5 & 53.6 & 81.6 & 72.1 & 88.0 & - & 95.4 & 88.0 & 88.8 & 78.2 \\
    TransReID\cite{he2021transreid} & 64.2 & 55.7 & 70.2 & 67.3 & 71.3 & 68.6 & 95.0 & 88.2 & 89.6 & 80.6\\
    FED\cite{wang2022feature} & 68.1 & 56.4 & \textbf{86.3} & 79.3 & 83.1 & 80.5 & 95.0 & 86.3 & 89.4 & 78.0\\
    DPM\cite{tan2022dynamic} & 69.9 & 60.7 & 82.7 & 76.6 & - & - & - & - & - & -\\
    DRL-Net\cite{jia2022learning} & - & - & - & - & - & - & 94.7 & 86.9 & 88.1  & 76.6 \\
    FRT\cite{xu2022frt} & 70.7 & 61.3 & 80.4 & 71.0 & \textbf{88.2} & - & 95.5 & 88.1 & 90.5 & \textbf{81.7}\\
    \midrule[0.5pt] 
    DDRN(Ours) & \textbf{71.3} & \textbf{62.4} & 84.7 & \textbf{80.7} & 83.3 & \textbf{80.6}& \textbf{95.8} & 88.4  & \textbf{90.8} & \textbf{81.7} \\
    \bottomrule[1pt]     
    
    \end{tabular}
}
\label{table1}  
\end{table}

\subsection{Comparison with State-of-the-art Methods}
We compare our DDRN with existing state-of-the-art (SOTA) person ReID approaches on three datasets: occluded, partial, and holistic.

\textbf{Results on Occluded Datasets:} We present a comprehensive comparison between our DDRN and existing SOTA person ReID methods on three challenging occluded datasets: Occluded-Duke\cite{miao2019pose}, Occluded-ReID\cite{zhuo2018occluded}, and Partial-ReID\cite{zheng2015partial}. Our evaluation includes popular methods such as \cite{sun2018beyond,zhong2020random,he2018deep,ge2018fd,he2018recognizing,he2019foreground,gao2020pose,miao2019pose,wang2020high,chen2021occlude,zhu2020identity,li2021diverse,yan2021occluded,he2021transreid,wang2022feature,tan2022dynamic,zang2021learning,somers2023body,xu2022frt}, among them, \cite{li2021diverse,he2021transreid,wang2022feature,tan2022dynamic,xu2022frt} is based on Transformer, and the rest are based on ResNet. We observe that Transformer-based models consistently outperform traditional convolutional neural network-based methods in occluded person ReID, showcasing the superior accuracy of Transformer-based approaches.

The comprehensive results are presented in Table \ref{table1}. By leveraging the Embedding Space, Orthogonal Loss, and HS-Arcface Loss, DDRN demonstrates significant improvements across all evaluation metrics, particularly on the challenging Occluded-Duke\cite{miao2019pose} dataset, We achieved an accuracy of Rank-1 71.3\% and mAP 62.4\%. Here, we surpass the current best-performing network by 0.6\% in Rank-1 accuracy and 1.1\% in mAP for occluded person ReID. On the Partial ReID\cite{zheng2015partial} dataset, our approach also outperforms the SOTA method by 0.1\% in mAP, achieving an overall accuracy 80.6\% (mAP) and accuracy 83.3\% (Rank-1). On the Occluded-ReID\cite{zhuo2018occluded} dataset, DDRN demonstrates excellent performance, achieving a 1.4\% improvement in mAP accuracy compared to the SOTA results. Overall, DDRN showcases its efficacy and superiority, particularly on the challenging Occluded-Duke dataset.

\textbf{Results on Holistic Datasets.} DDRN achieves excellent accuracy in occluded person ReID but also holistic person ReID. We evaluate our proposed model on the Market1501\cite{zheng2015scalable} and DukeMTMC-ReID\cite{zheng2017unlabeled} datasets and compare it with the previous SOTA approaches \cite{sun2018beyond,sun2019perceive,wang2018learning,miao2019pose,wang2020high,chen2021occlude,li2021diverse,he2021transreid,wang2022feature}.

The results in Table \ref{table1} reveal the superiority of DDRN. On DukeMTMC-reID, our model achieves SOTA with a Rank-1 accuracy of 90.8\% and an mAP of 81.7\%, exceeding the performance of the current SOTA method, FRT\cite{xu2022frt}, by 0.3\% in terms of mAP. On the Market-1501 dataset, DDRN demonstrates a 0.3\% improvement in both mAP and Rank-1 compared to FRT.

In summary, our DDRN demonstrates remarkable performance in both occluded and holistic person ReID tasks, showcasing its effectiveness and competitiveness against state-of-the-art methods across various challenging datasets.

\begin{table*}
\newcommand{\tabincell}[2]{\begin{tabular}{@{}#1@{}}#2\end{tabular}}
\centering  
\caption{Performance analysis of each component in DDRN.}  
\setlength{\tabcolsep}{1.5mm}{
    \begin{tabular}{ccccccccc}
    \toprule[1pt] 
    \multicolumn{6}{c}{Occluded-Duke}  \\
    \midrule[0.5pt]  
    Index & \tabincell{c}{Embedding Space} & \tabincell{c}{Orthogonal Loss} & \tabincell{c}{HS-Arcface} & R@1 & mAP\\ 
    \midrule[0.5pt]  
    0 & × & × & × & 62.5 & 55.2 \\
    1 & \checkmark & × & × & 65.7 & 59.0 \\
    2 & × & × & \checkmark & 66.5 & 56.6 \\
    3 & \checkmark & \checkmark & × & 67.1 & 60.0 \\
    4 & \checkmark & \checkmark & \checkmark & 71.3 & 62.4 \\
    \bottomrule[1pt]     
    \end{tabular}
}
\label{table3}  
\end{table*}

\subsection{Ablation Study}
The Occluded-Duke dataset's ablation study crucially assesses DDRN's effectiveness. In Table \ref{table3}, DDRN significantly outperforms the baseline, with an 8.8\% increase in Rank-1 accuracy and a remarkable 7.2\% improvement in mAP. From Table \ref{table3}, it can be observed that the utilization of Embedding Space contributes significantly to the improvement in mAP accuracy, achieving a notable increase of 3.8\%. In summary, the ablation study demonstrates the crucial contributions of all proposed components, culminating in DDRN's impressive performance.  DDRN's success results from the thoughtful combination of each component, offering a robust solution for occluded person ReID.


\begin{figure*}
    \centering
    \includegraphics[width=1.0\linewidth]{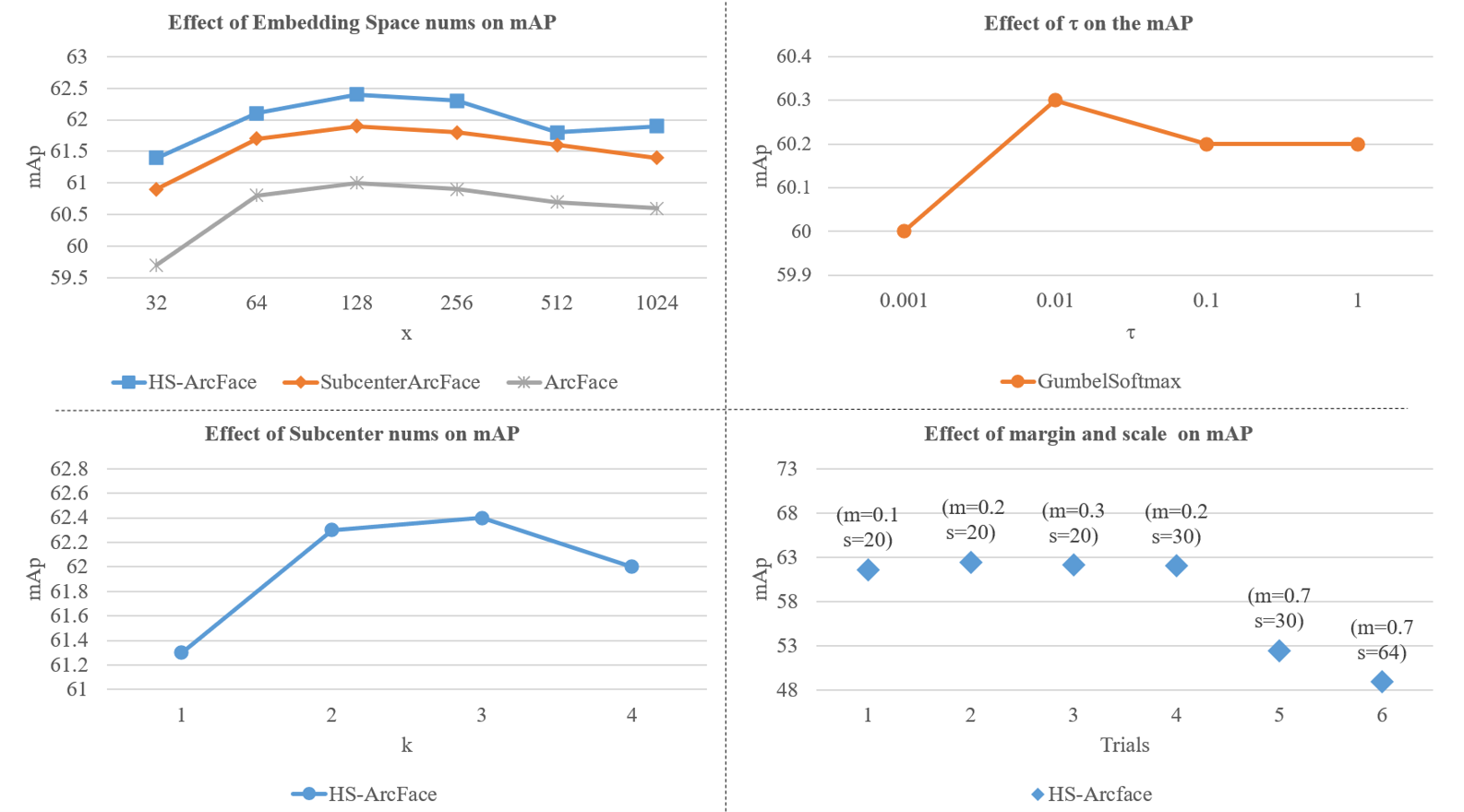}
    \caption{The selection of hyperparameters during the training process of DDRN. Here, we present several important parameters and their impact on the experimental accuracy, including the number $x$ of vectors in the Embedding Space, the value of $'\tau'$ in GumbelSoftmax, the number $k$ of sub-centers in HS-Arcface, and the influence of $'margin'$ and $'scale'$ hyperparameters in HS-Arcface on mAP. Furthermore, we provide a comparison between HS-Arcface, SubCenterArcFace, and ArcFace, as well as a comparison between default sampling and GumbelSoftmax sampling, as indicated in the legend.}
    \label{fig:6}
\end{figure*}
\subsection{Hyperparameter Selection}

In Figure \ref{fig:6}, we provide a comprehensive visualization of the critical parameter impact in DDRN on the mAP metric for the Occ-Duke dataset. We compare mAP changes using ArcFace, SubcenterArcFace, and our HS-Arcface. Results reveal HS-Arcface's superiority across hyperparameter settings. Extensive experimentation shows HS-Arcface consistently outperforms mainstream loss functions, enhancing accuracy. Optimal performance on Occ-Duke involves selecting 128 vectors in the Embedding Space, a Gumbel Softmax temperature of 0.1, 3 subcenters for balanced generalization, and setting HS-Arcface margin and scale to 0.2 and 20. These selections, a result of meticulous analysis, enable DDRN to achieve state-of-the-art performance, emphasizing its effectiveness in handling occlusion and background interference challenges in person ReID.

\begin{figure*}[!ht]
\centering
\includegraphics[width=1.0\linewidth]{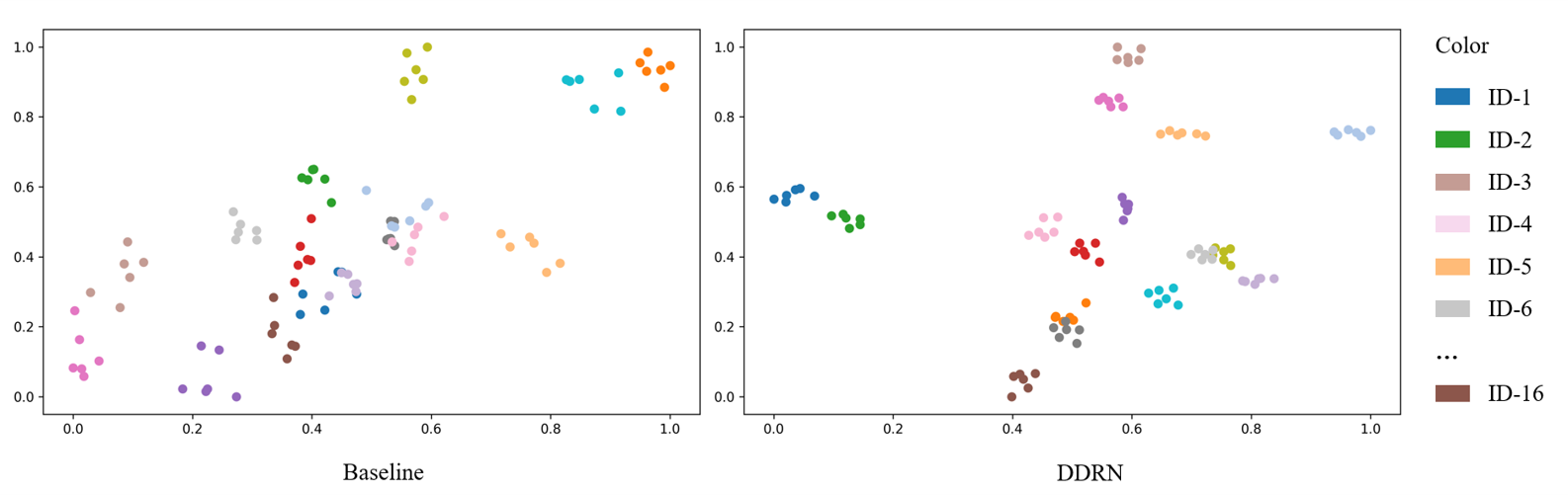}
\caption{On the Occluded-DukeMTMC dataset, t-SNE\cite{hinton2008visualizing}visualization of image features. Different colors represent different identities. We have selected 16 types of identities with similar occlusions and backgrounds. It is obvious that our proposed DDRN method can better cope with the interference of occlusion and background than the baseline method.}
\label{fig:7}
\end{figure*}
 
\subsection{Visualization of Feature Distribution}
To provide a more intuitive evaluation of our method, we used the t-SNE\cite{hinton2008visualizing} method to visualize the high-dimensional feature vectors output by the network, as shown in Fig.\ref{fig:7}. The Baseline struggles with occlusions, resulting in inaccuracies. DDRN, however, demonstrates remarkable resilience, reducing occlusion effects for more accurate retrievals. DDRN also shows robustness against background interference, filtering irrelevant information and focusing on discriminative features. This comparison highlights DDRN's efficacy in addressing occlusion and background challenges, positioning it as a promising solution for practical occluded person ReID applications.

\section{Conclusion}
This paper presents DDRN, a novel approach for occluded person ReID that diverges from conventional models. DDRN employs a generative model, predicting data distribution and reconstructing features to mitigate occlusion interference. Avoiding direct mean and variance prediction, we adopt an implicit approach—learning the Embedding Space. Gumbel Softmax enhances robustness during replacements, and Orthogonal Loss minimizes redundancy, improving generalization. HS-Arcface addresses severe occlusions, demonstrating robustness. DDRN pioneers data distribution-based feature reconstruction in occluded person ReID, advancing the field.

\clearpage  

%
%
\bibliographystyle{splncs04}
\bibliography{main}

\begin{thebibliography}{10}
\providecommand{\url}[1]{\texttt{#1}}
\providecommand{\urlprefix}{URL }
\providecommand{\doi}[1]{https://doi.org/#1}

\bibitem{bengio2013estimating}
Bengio, Y., L{\'e}onard, N., Courville, A.: Estimating or propagating gradients through stochastic neurons for conditional computation. arXiv preprint arXiv:1308.3432  (2013)

\bibitem{chang2018multi}
Chang, X., Hospedales, T.M., Xiang, T.: Multi-level factorisation net for person re-identification. In: Proceedings of the IEEE conference on computer vision and pattern recognition. pp. 2109--2118 (2018)

\bibitem{chen2021occlude}
Chen, P., Liu, W., Dai, P., Liu, J., Ye, Q., Xu, M., Chen, Q., Ji, R.: Occlude them all: Occlusion-aware attention network for occluded person re-id. In: Proceedings of the IEEE/CVF International Conference on Computer Vision. pp. 11833--11842 (2021)

\bibitem{deng2020sub}
Deng, J., Guo, J., Liu, T., Gong, M., Zafeiriou, S.: Sub-center arcface: Boosting face recognition by large-scale noisy web faces. In: Computer Vision--ECCV 2020: 16th European Conference, Glasgow, UK, August 23--28, 2020, Proceedings, Part XI 16. pp. 741--757. Springer (2020)

\bibitem{dosovitskiy2020image}
Dosovitskiy, A., Beyer, L., Kolesnikov, A., Weissenborn, D., Zhai, X., Unterthiner, T., Dehghani, M., Minderer, M., Heigold, G., Gelly, S., et~al.: An image is worth 16x16 words: Transformers for image recognition at scale. arXiv preprint arXiv:2010.11929  (2020)

\bibitem{eom2019learning}
Eom, C., Ham, B.: Learning disentangled representation for robust person re-identification. Advances in neural information processing systems  \textbf{32} (2019)

\bibitem{gao2020pose}
Gao, S., Wang, J., Lu, H., Liu, Z.: Pose-guided visible part matching for occluded person reid. In: Proceedings of the IEEE/CVF Conference on Computer Vision and Pattern Recognition. pp. 11744--11752 (2020)

\bibitem{ge2018fd}
Ge, Y., Li, Z., Zhao, H., Yin, G., Yi, S., Wang, X., et~al.: Fd-gan: Pose-guided feature distilling gan for robust person re-identification. Advances in neural information processing systems  \textbf{31} (2018)

\bibitem{goodfellow2020generative}
Goodfellow, I., Pouget-Abadie, J., Mirza, M., Xu, B., Warde-Farley, D., Ozair, S., Courville, A., Bengio, Y.: Generative adversarial networks. Communications of the ACM  \textbf{63}(11),  139--144 (2020)

\bibitem{guo2022attention}
Guo, M.H., Xu, T.X., Liu, J.J., Liu, Z.N., Jiang, P.T., Mu, T.J., Zhang, S.H., Martin, R.R., Cheng, M.M., Hu, S.M.: Attention mechanisms in computer vision: A survey. Computational Visual Media pp. 1--38 (2022)

\bibitem{he2018deep}
He, L., Liang, J., Li, H., Sun, Z.: Deep spatial feature reconstruction for partial person re-identification: Alignment-free approach. In: Proceedings of the IEEE conference on computer vision and pattern recognition. pp. 7073--7082 (2018)

\bibitem{he2018recognizing}
He, L., Sun, Z., Zhu, Y., Wang, Y.: Recognizing partial biometric patterns. arXiv preprint arXiv:1810.07399  (2018)

\bibitem{he2019foreground}
He, L., Wang, Y., Liu, W., Zhao, H., Sun, Z., Feng, J.: Foreground-aware pyramid reconstruction for alignment-free occluded person re-identification. In: Proceedings of the IEEE/CVF international conference on computer vision. pp. 8450--8459 (2019)

\bibitem{he2021transreid}
He, S., Luo, H., Wang, P., Wang, F., Li, H., Jiang, W.: Transreid: Transformer-based object re-identification. In: Proceedings of the IEEE/CVF international conference on computer vision. pp. 15013--15022 (2021)

\bibitem{hinton2008visualizing}
Hinton, G., van~der Maaten, L.: Visualizing data using t-sne journal of machine learning research  (2008)

\bibitem{huang2020human}
Huang, H., Chen, X., Huang, K.: Human parsing based alignment with multi-task learning for occluded person re-identification. In: 2020 IEEE International Conference on Multimedia and Expo (ICME). pp.~1--6. IEEE (2020)

\bibitem{jang2016categorical}
Jang, E., Gu, S., Poole, B.: Categorical reparameterization with gumbel-softmax. arXiv preprint arXiv:1611.01144  (2016)

\bibitem{jia2021matching}
Jia, M., Cheng, X., Zhai, Y., Lu, S., Ma, S., Tian, Y., Zhang, J.: Matching on sets: Conquer occluded person re-identification without alignment. In: National Conference on Artificial Intelligence (2021)

\bibitem{jia2022learning}
Jia, M., Cheng, X., Lu, S., Zhang, J.: Learning disentangled representation implicitly via transformer for occluded person re-identification. IEEE Transactions on Multimedia  \textbf{25},  1294--1305 (2022)

\bibitem{kalayeh2018human}
Kalayeh, M.M., Basaran, E., G{\"o}kmen, M., Kamasak, M.E., Shah, M.: Human semantic parsing for person re-identification. In: Proceedings of the IEEE conference on computer vision and pattern recognition. pp. 1062--1071 (2018)

\bibitem{kingma2013auto}
Kingma, D.P., Welling, M.: Auto-encoding variational bayes. arXiv preprint arXiv:1312.6114  (2013)

\bibitem{li2017learning}
Li, D., Chen, X., Zhang, Z., Huang, K.: Learning deep context-aware features over body and latent parts for person re-identification. In: Proceedings of the IEEE conference on computer vision and pattern recognition. pp. 384--393 (2017)

\bibitem{li2021combined}
Li, H., Wu, G., Zheng, W.S.: Combined depth space based architecture search for person re-identification. In: Proceedings of the IEEE/CVF Conference on Computer Vision and Pattern Recognition. pp. 6729--6738 (2021)

\bibitem{li2014deepreid}
Li, W., Zhao, R., Xiao, T., Wang, X.: Deepreid: Deep filter pairing neural network for person re-identification. In: Proceedings of the IEEE conference on computer vision and pattern recognition. pp. 152--159 (2014)

\bibitem{li2018harmonious}
Li, W., Zhu, X., Gong, S.: Harmonious attention network for person re-identification. In: Proceedings of the IEEE conference on computer vision and pattern recognition. pp. 2285--2294 (2018)

\bibitem{li2021diverse}
Li, Y., He, J., Zhang, T., Liu, X., Zhang, Y., Wu, F.: Diverse part discovery: Occluded person re-identification with part-aware transformer. In: Proceedings of the IEEE/CVF Conference on Computer Vision and Pattern Recognition. pp. 2898--2907 (2021)

\bibitem{luo2019bag}
Luo, H., Gu, Y., Liao, X., Lai, S., Jiang, W.: Bag of tricks and a strong baseline for deep person re-identification. In: Proceedings of the IEEE/CVF conference on computer vision and pattern recognition workshops. pp.~0--0 (2019)

\bibitem{luo2020stnreid}
Luo, H., Jiang, W., Fan, X., Zhang, C.: Stnreid: Deep convolutional networks with pairwise spatial transformer networks for partial person re-identification. IEEE Transactions on Multimedia  \textbf{22}(11),  2905--2913 (2020)

\bibitem{2021MaPirt}
Ma, Z., Zhao, Y., Li, J.: Pose-guided inter- and intra-part relational transformer for occluded person re-identification. In: Proceedings of the 29th ACM International Conference on Multimedia. p. 1487–1496. MM '21, Association for Computing Machinery, New York, NY, USA (2021)

\bibitem{miao2019pose}
Miao, J., Wu, Y., Liu, P., Ding, Y., Yang, Y.: Pose-guided feature alignment for occluded person re-identification. In: Proceedings of the IEEE/CVF international conference on computer vision. pp. 542--551 (2019)

\bibitem{rao2021counterfactual}
Rao, Y., Chen, G., Lu, J., Zhou, J.: Counterfactual attention learning for fine-grained visual categorization and re-identification. In: Proceedings of the IEEE/CVF International Conference on Computer Vision. pp. 1025--1034 (2021)

\bibitem{ren2020semantic}
Ren, X., Zhang, D., Bao, X.: Semantic-guided shared feature alignment for occluded person re-identification. In: Asian Conference on Machine Learning. pp. 17--32. PMLR (2020)

\bibitem{sarfraz2018pose}
Sarfraz, M.S., Schumann, A., Eberle, A., Stiefelhagen, R.: A pose-sensitive embedding for person re-identification with expanded cross neighborhood re-ranking. In: Proceedings of the IEEE conference on computer vision and pattern recognition. pp. 420--429 (2018)

\bibitem{shen2018person}
Shen, Y., Li, H., Yi, S., Chen, D., Wang, X.: Person re-identification with deep similarity-guided graph neural network. In: Proceedings of the European conference on computer vision (ECCV). pp. 486--504 (2018)

\bibitem{somers2023body}
Somers, V., De~Vleeschouwer, C., Alahi, A.: Body part-based representation learning for occluded person re-identification. In: Proceedings of the IEEE/CVF Winter Conference on Applications of Computer Vision. pp. 1613--1623 (2023)

\bibitem{su2017pose}
Su, C., Li, J., Zhang, S., Xing, J., Gao, W., Tian, Q.: Pose-driven deep convolutional model for person re-identification. In: Proceedings of the IEEE international conference on computer vision. pp. 3960--3969 (2017)

\bibitem{sun2019perceive}
Sun, Y., Xu, Q., Li, Y., Zhang, C., Li, Y., Wang, S., Sun, J.: Perceive where to focus: Learning visibility-aware part-level features for partial person re-identification. In: Proceedings of the IEEE/CVF conference on computer vision and pattern recognition. pp. 393--402 (2019)

\bibitem{sun2018beyond}
Sun, Y., Zheng, L., Yang, Y., Tian, Q., Wang, S.: Beyond part models: Person retrieval with refined part pooling (and a strong convolutional baseline). In: Proceedings of the European conference on computer vision (ECCV). pp. 480--496 (2018)

\bibitem{tan2022mhsa}
Tan, H., Liu, X., Yin, B., Li, X.: Mhsa-net: Multihead self-attention network for occluded person re-identification. IEEE Transactions on Neural Networks and Learning Systems  (2022)

\bibitem{tan2022dynamic}
Tan, L., Dai, P., Ji, R., Wu, Y.: Dynamic prototype mask for occluded person re-identification. In: Proceedings of the 30th ACM International Conference on Multimedia. pp. 531--540 (2022)

\bibitem{tay2019aanet}
Tay, C.P., Roy, S., Yap, K.H.: Aanet: Attribute attention network for person re-identifications. In: Proceedings of the IEEE/CVF conference on computer vision and pattern recognition. pp. 7134--7143 (2019)

\bibitem{van2017neural}
Van Den~Oord, A., Vinyals, O., et~al.: Neural discrete representation learning. Advances in neural information processing systems  \textbf{30} (2017)

\bibitem{wang2020high}
Wang, G., Yang, S., Liu, H., Wang, Z., Yang, Y., Wang, S., Yu, G., Zhou, E., Sun, J.: High-order information matters: Learning relation and topology for occluded person re-identification. In: Proceedings of the IEEE/CVF conference on computer vision and pattern recognition. pp. 6449--6458 (2020)

\bibitem{wang2018learning}
Wang, G., Yuan, Y., Chen, X., Li, J., Zhou, X.: Learning discriminative features with multiple granularities for person re-identification. In: Proceedings of the 26th ACM international conference on Multimedia. pp. 274--282 (2018)

\bibitem{wang2018transferable}
Wang, J., Zhu, X., Gong, S., Li, W.: Transferable joint attribute-identity deep learning for unsupervised person re-identification. In: Proceedings of the IEEE conference on computer vision and pattern recognition. pp. 2275--2284 (2018)

\bibitem{wang2022feature}
Wang, Z., Zhu, F., Tang, S., Zhao, R., He, L., Song, J.: Feature erasing and diffusion network for occluded person re-identification. In: Proceedings of the IEEE/CVF Conference on Computer Vision and Pattern Recognition. pp. 4754--4763 (2022)

\bibitem{wei2018person}
Wei, L., Zhang, S., Gao, W., Tian, Q.: Person transfer gan to bridge domain gap for person re-identification. In: Proceedings of the IEEE conference on computer vision and pattern recognition. pp. 79--88 (2018)

\bibitem{wu2016personnet}
Wu, L., Shen, C., Hengel, A.v.d.: Personnet: Person re-identification with deep convolutional neural networks. arXiv preprint arXiv:1601.07255  (2016)

\bibitem{xu2022frt}
Xu, B., He, L., Liang, J., Sun, Z.: Learning feature recovery transformer for occluded person re-identification. IEEE Transactions on Image Processing  \textbf{31},  4651--4662 (2022)

\bibitem{yan2021beyond}
Yan, C., Pang, G., Bai, X., Liu, C., Ning, X., Gu, L., Zhou, J.: Beyond triplet loss: person re-identification with fine-grained difference-aware pairwise loss. IEEE Transactions on Multimedia  \textbf{24},  1665--1677 (2021)

\bibitem{yan2021occluded}
Yan, C., Pang, G., Jiao, J., Bai, X., Feng, X., Shen, C.: Occluded person re-identification with single-scale global representations. In: Proceedings of the IEEE/CVF International Conference on Computer Vision. pp. 11875--11884 (2021)

\bibitem{yang2019towards}
Yang, W., Huang, H., Zhang, Z., Chen, X., Huang, K., Zhang, S.: Towards rich feature discovery with class activation maps augmentation for person re-identification. In: Proceedings of the IEEE/CVF conference on computer vision and pattern recognition. pp. 1389--1398 (2019)

\bibitem{ye2021channel}
Ye, M., Ruan, W., Du, B., Shou, M.Z.: Channel augmented joint learning for visible-infrared recognition. In: Proceedings of the IEEE/CVF International Conference on Computer Vision. pp. 13567--13576 (2021)

\bibitem{zang2021learning}
Zang, X., Li, G., Gao, W., Shu, X.: Learning to disentangle scenes for person re-identification. Image and Vision Computing  \textbf{116},  104330 (2021)

\bibitem{zhai2020ad}
Zhai, Y., Lu, S., Ye, Q., Shan, X., Chen, J., Ji, R., Tian, Y.: Ad-cluster: Augmented discriminative clustering for domain adaptive person re-identification. In: Proceedings of the IEEE/CVF conference on computer vision and pattern recognition. pp. 9021--9030 (2020)

\bibitem{zhang2020semantic}
Zhang, X., Yan, Y., Xue, J.H., Hua, Y., Wang, H.: Semantic-aware occlusion-robust network for occluded person re-identification. IEEE Transactions on Circuits and Systems for Video Technology  \textbf{31}(7),  2764--2778 (2020)

\bibitem{zhang2017alignedreid}
Zhang, X., Luo, H., Fan, X., Xiang, W., Sun, Y., Xiao, Q., Jiang, W., Zhang, C., Sun, J.: Alignedreid: Surpassing human-level performance in person re-identification. arXiv preprint arXiv:1711.08184  (2017)

\bibitem{zhang2020relation}
Zhang, Z., Lan, C., Zeng, W., Jin, X., Chen, Z.: Relation-aware global attention for person re-identification. In: Proceedings of the ieee/cvf conference on computer vision and pattern recognition. pp. 3186--3195 (2020)

\bibitem{zheng2019pyramidal}
Zheng, F., Deng, C., Sun, X., Jiang, X., Guo, X., Yu, Z., Huang, F., Ji, R.: Pyramidal person re-identification via multi-loss dynamic training. In: Proceedings of the IEEE/CVF conference on computer vision and pattern recognition. pp. 8514--8522 (2019)

\bibitem{zheng2015scalable}
Zheng, L., Shen, L., Tian, L., Wang, S., Wang, J., Tian, Q.: Scalable person re-identification: A benchmark. In: Proceedings of the IEEE international conference on computer vision. pp. 1116--1124 (2015)

\bibitem{zheng2015partial}
Zheng, W.S., Li, X., Xiang, T., Liao, S., Lai, J., Gong, S.: Partial person re-identification. In: Proceedings of the IEEE International Conference on Computer Vision. pp. 4678--4686 (2015)

\bibitem{zheng2017unlabeled}
Zheng, Z., Zheng, L., Yang, Y.: Unlabeled samples generated by gan improve the person re-identification baseline in vitro. In: Proceedings of the IEEE international conference on computer vision. pp. 3754--3762 (2017)

\bibitem{zhong2020random}
Zhong, Z., Zheng, L., Kang, G., Li, S., Yang, Y.: Random erasing data augmentation. In: Proceedings of the AAAI conference on artificial intelligence. vol.~34, pp. 13001--13008 (2020)

\bibitem{zhu2020identity}
Zhu, K., Guo, H., Liu, Z., Tang, M., Wang, J.: Identity-guided human semantic parsing for person re-identification. In: Computer Vision--ECCV 2020: 16th European Conference, Glasgow, UK, August 23--28, 2020, Proceedings, Part III 16. pp. 346--363. Springer (2020)

\bibitem{zhu2023aaformer}
Zhu, K., Guo, H., Zhang, S., Wang, Y., Liu, J., Wang, J., Tang, M.: Aaformer: Auto-aligned transformer for person re-identification. IEEE Transactions on Neural Networks and Learning Systems  (2023)

\bibitem{zhuo2018occluded}
Zhuo, J., Chen, Z., Lai, J., Wang, G.: Occluded person re-identification. In: 2018 IEEE International Conference on Multimedia and Expo (ICME). pp.~1--6. IEEE (2018)

\end{thebibliography}
\end{document}